\documentclass[11pt]{article}

\usepackage[preprint]{acl} 

\usepackage{times}
\usepackage{latexsym}
\usepackage{amsmath}
\usepackage{amssymb}

\usepackage{algorithm}
\usepackage{algpseudocode}
\usepackage{multirow}
\usepackage[table]{xcolor}

\usepackage{booktabs}
\usepackage{newfloat}
\usepackage{listings}
\DeclareCaptionStyle{ruled}{labelfont=normalfont,labelsep=colon,strut=off} 
\floatstyle{ruled}
\newfloat{listing}{tb}{lst}{}
\floatname{listing}{Listing}
\usepackage[T1]{fontenc}

\usepackage[utf8]{inputenc}

\usepackage{microtype}

\usepackage{inconsolata}

\usepackage{graphicx}

\title{Learning What to Keep: Transfer-Aware Personal Skill Learning from User Feedback}

\author{
  \textbf{Li Wang}$^{1}$ \quad
  \textbf{Chen Zhang}$^{1}$ \quad
  \textbf{Yirong Yang}$^{1}$ \quad
  \textbf{Xingyu Chen}$^{2}$ \\
  \textbf{Jinkun Zhao}$^{1}$ \quad
  \textbf{Tianhao Peng}$^{3}$\thanks{Corresponding author: Tianhao Peng, email: pengtianhao@buaa.edu.cn, Wenjun Wu, email: wwj09315@buaa.edu.cn} \quad
  \textbf{Wenjun Wu}$^{1,4,5}$\footnotemark[1] \\
  \\
  $^{1}$Beihang University \\
  $^{2}$Chengdu University of Information Technology \\
  $^{3}$Nanyang Technological University \\
  $^{4}$Hangzhou International Innovation Institute \\
  $^{5}$Beijing Advanced Innovation Center for Future Blockchain and Privacy Computing \\
}

\usepackage[most]{tcolorbox}
\usepackage{flafter}
\usepackage{placeins}

\definecolor{promptbar}{HTML}{3F3D3D}
\definecolor{casegray}{HTML}{4A4A4A}
\definecolor{caseblue}{HTML}{355C7D}
\definecolor{casegreen}{HTML}{3F7D62}
\definecolor{casegold}{HTML}{9A6A1F}

\newtcblisting{promptbox}[1]{
  enhanced,
  listing only,
  colback=white,
  colframe=black,
  coltitle=white,
  colbacktitle=promptbar,
  title={#1},
  fonttitle=\sffamily\footnotesize,
  boxrule=0.8pt,
  arc=2mm,
  left=2mm,
  right=2mm,
  top=1mm,
  bottom=1mm,
  toptitle=1mm,
  bottomtitle=1mm,
  listing options={
    basicstyle=\sffamily\fontsize{7.2pt}{8.6pt}\selectfont,
    numbers=none,
    xleftmargin=0pt,
    breaklines=true,
    breakatwhitespace=false,
    breakindent=0pt,
    breakautoindent=false,
    columns=fullflexible,
    keepspaces=true,
    showstringspaces=false
  }
}

\newtcblisting{promptcolorbox}[2]{
  enhanced,
  listing only,
  colback=#2!3!white,
  colframe=#2,
  coltitle=white,
  colbacktitle=#2,
  title={#1},
  fonttitle=\sffamily\bfseries\footnotesize,
  boxrule=0.8pt,
  arc=2mm,
  left=2mm,
  right=2mm,
  top=1mm,
  bottom=1mm,
  toptitle=1mm,
  bottomtitle=1mm,
  listing options={
    basicstyle=\sffamily\fontsize{7.2pt}{8.6pt}\selectfont,
    numbers=none,
    xleftmargin=0pt,
    breaklines=true,
    breakatwhitespace=false,
    breakindent=0pt,
    breakautoindent=false,
    columns=fullflexible,
    keepspaces=true,
    showstringspaces=false
  }
}

\newtcolorbox{casebox}[2]{
  enhanced,
  colback=#2!4!white,
  colframe=#2,
  coltitle=white,
  colbacktitle=#2,
  title={#1},
  fonttitle=\sffamily\bfseries\footnotesize,
  boxrule=0.7pt,
  arc=1.5mm,
  left=1.5mm,
  right=1.5mm,
  top=1mm,
  bottom=1mm,
  toptitle=1mm,
  bottomtitle=1mm,
  before skip=5pt,
  after skip=5pt
}

\begin{document}
\maketitle
\begin{abstract}
User feedback enables LLMs to refine reusable personalized skills, but often mixes persistent preferences with request-specific corrections. Indiscriminately retaining both risks negative transfer, motivating selective retention at the level of individual requirements. We frame this challenge as selective transfer over atomic rules and propose \textsc{SkillGrain}, a framework for learning reusable personalized skills from user feedback. \textsc{SkillGrain} decomposes feedback into independently actionable rules, predicts their transferability, and retains only those suitable for future requests. To learn these decisions, we combine rubric-guided skill rewards with fine-grained credit assignment that separately evaluates rule content and retention decisions. We further jointly optimize skill induction and skill-conditioned generation through reinforcement learning, ensuring that learned skills support downstream personalization. We construct a benchmark spanning five personalized writing tasks to evaluate cross-request generalization. Experiments across three base models in both in-domain and out-of-domain settings show that \textsc{SkillGrain} achieves the best overall average performance among competing methods. Our code is available at \url{https://anonymous.4open.science/r/SkillGrain-C210}.
\end{abstract}

\section{Introduction}
\begin{figure}[t]
    \centering
    \includegraphics[width=\linewidth]{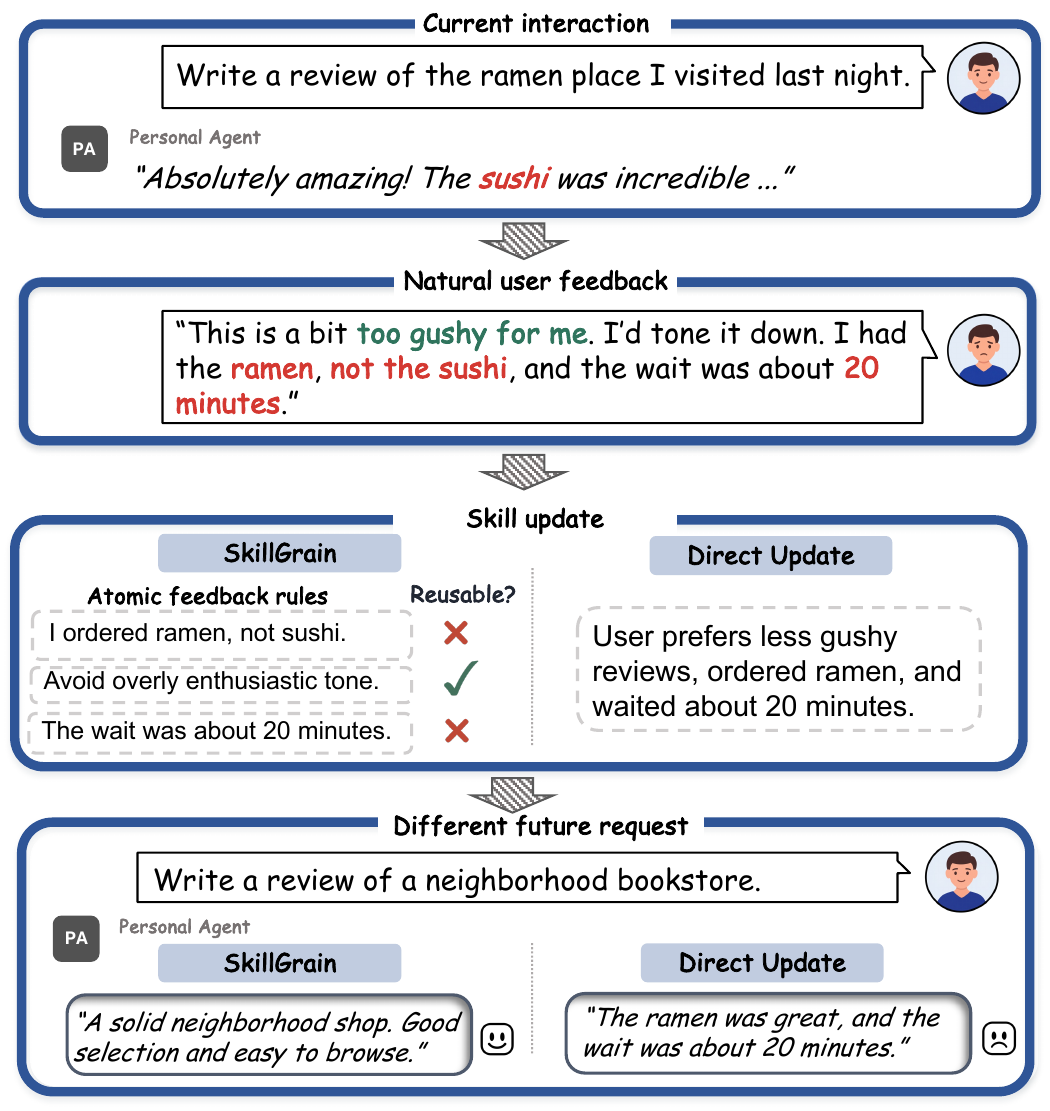}
    \caption{User feedback contains a mixture of transferable preferences and task-specific details. SkillGrain learns a user-specific skill from feedback to enable consistent personalized responses.}
    \label{fig:intro}
\end{figure}

Reusable skills provide Large Language Models (LLMs) with a mechanism to accumulate interaction experience and continuously improve future performance. In personalized scenarios, models can update reusable skills from user feedback to better adapt to individual preferences. However, user feedback may contain information with different scopes of applicability: as shown in Figure~\ref{fig:intro}, some feedback reflects transferable preferences that generalize across future requests, while other feedback only addresses requirements specific to the current interaction. For instance, feedback on writing style or organization can capture a user's persistent preference, whereas instructions involving specific entities or facts are often tied to the current request. Incorporating such heterogeneous information indiscriminately into persistent skills may introduce inappropriate constraints and lead to negative transfer. Therefore, effective skill learning requires models to identify the transferability of individual feedback components and selectively retain information that can generalize across future requests.

Prior work has explored experience reuse and personalization from different perspectives. Skill-evolution methods distill execution experience into reusable procedures and validate their effectiveness through task performance \cite{lin2026muse,tang2026wikiskill}. However, evaluating a skill as a whole does not directly reveal which individual feedback requirements should be retained. History-based personalization models user preferences through memories or summaries \cite{zhang2025prime,liu2026text}, but does not explicitly supervise whether each new correction should become a persistent preference. Scope-aware methods distinguish stable from situational preferences or restrict adaptation to relevant contexts \cite{gao2026beyond,stephan2024rlvf}, focusing on when preferences apply rather than which requirements within mixed feedback should be retained. More directly, correction-driven systems filter transient requirements or maintain rules with applicability conditions \cite{yang2026autoskill,zhou2026getting}, yet do not separately optimize the accuracy of rule extraction and retention decisions. Despite these advances, learning from feedback that mixes transferable preferences with request-specific corrections remains challenging. This raises a key question: can models learn to identify and selectively retain transferable preferences from such mixed feedback to better personalize future responses?

In this work, we formulate personalized skill learning as a selective transfer problem and introduce \textsc{SkillGrain}. Our framework decomposes mixed feedback into atomic rules and predicts their transferability, retaining reusable preferences while excluding request-specific corrections from persistent skills. To train these decisions, we combine rubric-guided skill rewards with fine-grained credit assignment, providing separate learning signals for rule content and transferability. We further jointly optimize skill induction and skill-conditioned generation through reinforcement learning, using distinct rewards for skill updates and downstream responses. To evaluate our approach, we construct a benchmark spanning five personalized writing tasks. Experiments across three base models in both in-domain and out-of-domain settings show that \textsc{SkillGrain} achieves the highest average score on each model. Our main contributions are as follows:
\begin{itemize}
    \item \textbf{Feedback-Driven Personalization as Selective Transfer.}
    We formulate personalized skill learning as selective transfer and construct a five-task writing benchmark that tests whether feedback-derived skills preserve reusable preferences without carrying over request-specific information.

    \item \textbf{Transfer-Aware Skill Induction and Credit Assignment.}
    We introduce \textsc{SkillGrain}, which pairs atomic rules with learned transferability decisions and assigns separate credit to rule content and retention, jointly optimizing skill induction and downstream application.

    \item \textbf{Generalization Across Models and Domains.}
    We evaluate \textsc{SkillGrain} across three base models in both in-domain and out-of-domain settings, achieving the highest overall average score on each model and demonstrating the effectiveness of our approach.
\end{itemize}

\section{Related work}
\subsection{Skill Evolution in LLM Agents}Recent work on LLM agents has increasingly explored how interaction experience can be distilled into reusable knowledge. Early approaches summarize task outcomes as verbal reflections or extract reusable lessons from multiple trajectories~\cite{shinn2023reflexion,zhao2024expel}, while subsequent methods induce and retrieve workflows or skills to guide future tasks~\cite{wang2024agent}. More recent work further studies continual skill evolution, allowing skills to be created, evaluated, revised, and maintained as new experience is collected~\cite{lin2026muse,tang2026wikiskill,he2026skillcommit,tang2026memory}. These methods focus on accumulating and refining reusable task knowledge from interaction experience. Our work extends this direction to personalized skill learning, where explicit user feedback must be selectively incorporated into skills that are reused across subsequent requests.

\subsection{Large Language Model Personalization}
LLM personalization commonly exploits interaction history through user representations or preference learning. One line of work aggregates histories or infers compact personas so that generation can be explicitly conditioned on user attributes~\cite{liu2025llms+,balepur2025whose}, whereas another learns shared and user-specific preference signals across users, for example through collaborative learning or preference calibration, to better adapt to sparse or unseen users~\cite{choi2025copl,ma2026personalizing}. Research on long-horizon personalization further organizes user experiences with episodic and semantic memories~\cite{zhang2025prime}, combines short-term deviations with long-term trends~\cite{sun2026preference}, and distinguishes stable preferences from context-dependent ones~\cite{gao2026beyond}; recent personalized agents also couple memory updating with action selection~\cite{zhang2026personaagent}. Taken together, these approaches primarily model personalization by extracting and maintaining user-specific information from accumulated interactions. We complement this line of research by distilling user feedback into reusable personalized skills.

\section{Task and Benchmark Construction}
\label{sec:benchmark}

\subsection{Task Formulation}
\label{sec:task_formulation}

We formulate personalized skill learning as a sequential interaction process between a user and an LLM. For each user $u$, let $H_u$ denote a fixed set of historical interactions and $s_t$ the personalized skill maintained at interaction step $t$. Given a user request $x_t$, the model generates a response
\begin{equation}
y_t \sim \pi_{\theta}
\bigl(\cdot \mid H_u, s_t, x_t\bigr).
\label{eq:response_generation}
\end{equation}
After observing user feedback $f_t$, the same model updates the personalized skill as
\begin{equation}
s_{t+1} \sim \pi_{\theta}
\bigl(\cdot \mid H_u, s_t, x_t, y_t, f_t\bigr).
\label{eq:skill_update_formulation}
\end{equation}
The updated skill $s_{t+1}$ is then used to guide subsequent requests from the same user. Our goal is to learn how interaction feedback can be distilled into reusable personalized skills that better capture user preferences while avoiding the transfer of interaction-specific information.

\subsection{Benchmark Construction}
\label{sec:benchmark_construction}

We construct a personalized skill-learning benchmark spanning five writing tasks: Topic Writing, Email, Yelp, Product Review, and Abstract. The first three are used for training and in-domain evaluation, while the remaining two are held out for out-of-domain evaluation. For each user, we select five historical writing examples and synthesize interaction queries conditioned on the user history and task setting. We use an LLM for query synthesis and apply automatic and LLM-based quality checks to remove duplicated, inconsistent, or low-quality instances. Detailed dataset statistics and construction procedures are provided in Appendix~\ref{appendix:benchmark_details}.

To simulate feedback containing both transferable preferences and request-specific requirements, we derive structured benchmark-side annotations
\begin{equation}
\mathcal{B}_{u,t}
=
\operatorname{Annotate}(H_u,x_t),
\label{eq:benchmark_annotation}
\end{equation}
which capture recurring writing preferences inferred from user history and interaction-specific content. We use a four-dimensional rubric to evaluate responses and guide feedback construction. The details of the response reward $R_{\mathrm{ans}}$ are provided in Appendix~\ref{appendix:rubric_scoring}. Given a response $y_t$, we derive structured revision targets through rubric-based diagnostics:
\begin{equation}
\mathcal{V}_t
=
\bigl[(v_m,\tau_m)\bigr]_{m=1}^{M_t}
=
\operatorname{Diagnose}
(x_t,y_t;\mathcal{B}_{u,t}),
\label{eq:revision_targets}
\end{equation}
where $v_m$ denotes an individual revision target and $\tau_m\in\{0,1\}$ indicates its transferability: $\tau_m=1$ for reusable preference revisions and $\tau_m=0$ for interaction-specific requirements. Finally, we verbalize the revision targets into natural-language feedback using a fixed LLM:
\begin{equation}
f_t=\mathcal{G}_{f}(\mathcal{V}_t).
\label{eq:feedback_generation}
\end{equation}
Further details are provided in Appendix~\ref{appendix:feedback_details}.

\begin{figure*}[t]
    \centering
    \includegraphics[width=\textwidth]{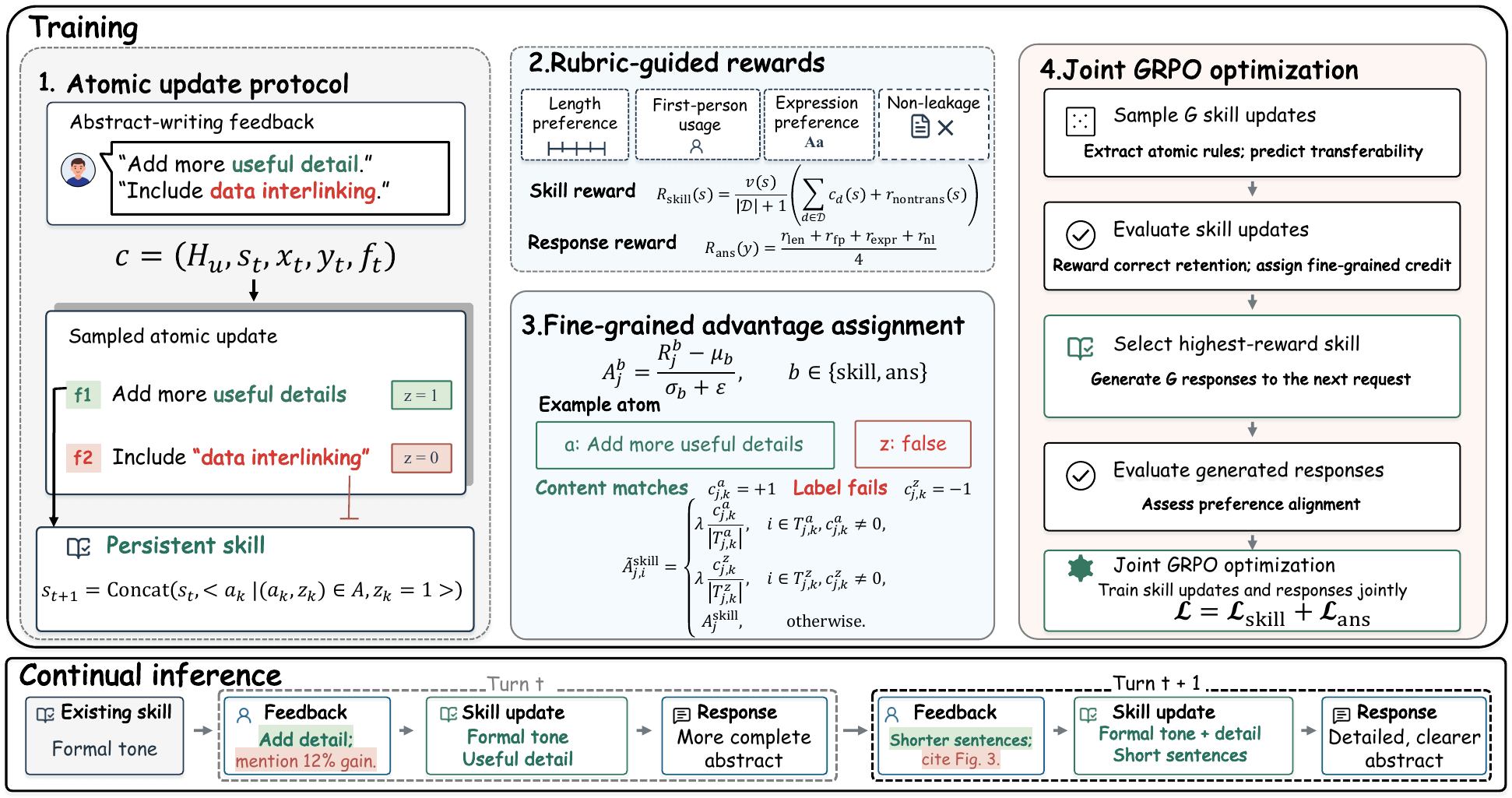}
    \caption{Overview of SkillGrain. We construct user-specific training instances consisting of an adaptation turn and subsequent requests. The model jointly optimizes skill-update and response trajectories using GRPO, with separate rubric-based rewards and independently normalized advantages. At inference time, the learned user skill is used to generate personalized responses.}
    \label{fig:method}
\end{figure*}

\section{Method}
\label{sec:method}
In this section, we present the design and training of \textsc{SkillGrain}, with an overview illustrated in Figure~\ref{fig:method}. We first introduce atomic skill induction in Sec.~\ref{sec:atomic_induction}, followed by rubric-guided reward design in Sec.~\ref{sec:rubric_rewards} and fine-grained advantage assignment in Sec.~\ref{sec:credit_assignment}. We then present the overall training pipeline and joint optimization objective in Sec.~\ref{sec:joint_optimization}. The detailed training procedure is provided in Appendix~\ref{appendix:training_algorithm}.

\subsection{Transfer-Aware Atomic Skill Induction}
\label{sec:atomic_induction}

User feedback may contain both transferable preferences and request-specific requirements. We therefore decompose feedback into atomic rules and predict the transferability of each rule individually. Following the task formulation in Sec.~\ref{sec:task_formulation}, we define the skill-update context as
\begin{equation}
c_t=(H_u,s_t,x_t,y_t,f_t).
\label{eq:interaction_context}
\end{equation}
Conditioned on $c_t$, the policy generates a sequence of atomic rules with their transferability decisions:
\begin{equation}
\mathcal{A}
=
\bigl[(a_k,z_k)\bigr]_{k=1}^{K}
\sim
\pi_{\theta}(\cdot\mid c_t),
\label{eq:atomic_induction}
\end{equation}
where $a_k$ represents an individual requirement expressed in the feedback, and $z_k\in\{0,1\}$ indicates whether the rule should be reused for subsequent requests. The model extracts both transferable and non-transferable requirements
and explicitly determines their applicability. Only rules predicted
to be transferable are appended to the existing personalized skill:
\begin{equation}
s_{t+1}
=
\operatorname{Concat}
\bigl(s_t,\langle a_k\mid (a_k,z_k)\in\mathcal A,\ z_k=1\rangle\bigr).
\label{eq:skill_composition}
\end{equation}
If no rule is predicted to be transferable, the existing skill remains unchanged.

\subsection{Rubric-Guided Rewards}
\label{sec:rubric_rewards}

Since correctly retaining transferable preferences and producing high-quality downstream responses are distinct objectives, we use separate rewards for personalized skill induction and skill-conditioned generation. For skill induction, the reward is designed to encourage two behaviors: preserving transferable preference information and preventing non-transferable information from being incorporated into the reusable skill. Let $\mathcal{D}$ denote the set of rubric dimensions that characterize transferable user preferences. For each dimension $d\in\mathcal{D}$, we define $c_d(s)\in\{0,1\}$ to indicate whether the candidate skill $s$ makes the correct update for that dimension. A correct decision may either incorporate a reusable preference revision identified from the current feedback or retain the existing rule when no revision is required. We separately evaluate whether the candidate skill incorrectly retains information whose applicability is limited to the current interaction. Let $n_{\mathrm{local}}(s)$ denote the number of distinct non-transferable elements incorporated into $s$, such as request-specific entities or one-off requirements. We define the corresponding reward as
\begin{equation}
r_{\mathrm{nontrans}}(s)
=
2^{-n_{\mathrm{local}}(s)}.
\label{eq:nontransfer_reward}
\end{equation}

This reward reaches its maximum when no interaction-specific information is retained and decreases as more non-transferable information is incorporated into the reusable skill. We further define $v(s)\in\{0,1\}$ to indicate whether the candidate makes a valid overall update decision. Specifically, the skill should be updated when the feedback contains reusable preference information and should remain unchanged otherwise. The overall skill reward is
\begin{equation}
R_{\mathrm{skill}}(s)
=
\frac{v(s)}{|\mathcal{D}|+1}
\left(
\sum_{d\in\mathcal{D}} c_d(s)
+
r_{\mathrm{nontrans}}(s)
\right).
\label{eq:skill_reward}
\end{equation}

In our benchmark, $\mathcal{D}$ corresponds to the verifiable preference dimensions defined by the rubric, while request-specific content and constraints are treated as non-transferable. $R_{\mathrm{skill}}$ directly evaluates the induced skill and optimizes skill-update trajectories, whereas $R_{\mathrm{ans}}$ defined in Appendix~\ref{appendix:rubric_scoring} supervises skill-conditioned response generation.

\subsection{Fine-Grained Advantage Assignment}
\label{sec:credit_assignment}

To enable finer-grained credit assignment, we introduce an advantage adjustment mechanism for rule content and transferability decisions. For each sampled skill-update trajectory $j$, we first compute its group-relative advantage as
\begin{equation}
A_j^{\mathrm{skill}}
=
\frac{
R_j^{\mathrm{skill}}-\mu_{\mathrm{skill}}
}{
\sigma_{\mathrm{skill}}+\epsilon
},
\label{eq:skill_advantage}
\end{equation}
where $\mu_{\mathrm{skill}}$ and $\sigma_{\mathrm{skill}}$ denote the mean and standard deviation of skill rewards within the sampled group. We then align each generated atom $(a_{j,k},z_{j,k})$ with a corresponding revision target $(v_m,\tau_m)\in\mathcal{V}_t$. For each reliably matched pair, we separately evaluate the correctness of the rule content and its transferability decision:
\begin{equation}
c_{j,k}^{a},\;c_{j,k}^{z}
\in\{-1,0,1\},
\label{eq:local_credit}
\end{equation}
where $c_{j,k}^{a}$ evaluates whether $a_{j,k}$ matches the target requirement $v_m$, and $c_{j,k}^{z}$ evaluates whether $z_{j,k}=\tau_m$. A value of $+1$ or $-1$ denotes a correct or incorrect decision, respectively, while $0$ denotes abstention when reliable alignment is unavailable. Let $T_{j,k}^{a}$ and $T_{j,k}^{z}$ denote the tokens corresponding to the rule-content and transferability fields. For reliably verified fields, we replace the trajectory-level advantage with a length-normalized local signal:
\begin{equation}
\widetilde{A}_{j,i}^{\mathrm{skill}}
=
\begin{cases}
\lambda \dfrac{c_{j,k}^{a}}{|T_{j,k}^{a}|},
& i\in T_{j,k}^{a},\;c_{j,k}^{a}\neq 0,\\[5pt]
\lambda \dfrac{c_{j,k}^{z}}{|T_{j,k}^{z}|},
& i\in T_{j,k}^{z},\;c_{j,k}^{z}\neq 0,\\[5pt]
A_j^{\mathrm{skill}},
& \text{otherwise}.
\end{cases}
\label{eq:local_advantage}
\end{equation}
Thus, rule content and transferability can receive different optimization signals within the same atom, while unverifiable tokens retain the trajectory-level advantage. For transferable targets omitted entirely from the generated atoms, we additionally assign an omission penalty to the sequence-closing decision. Detailed alignment and omission-penalty rules are provided in Appendix~\ref{appendix:credit_details}.

\subsection{Joint GRPO Training}
\label{sec:joint_optimization}

We jointly optimize skill updating and skill-conditioned generation using Group Relative Policy Optimization (GRPO)~\cite{shao2024deepseekmath}. Each training instance consists of an adaptation interaction followed by a subsequent request from the same user. For the skill-update stage, we sample $G$ candidate skill updates and evaluate them using $R_{\mathrm{skill}}$. Their group-relative advantages are computed using Eq.~\ref{eq:skill_advantage} and refined by the fine-grained signals in Eq.~\ref{eq:local_advantage}. We then select the highest-reward skill to generate $G$ responses to the subsequent training query. These responses are evaluated using $R_{\mathrm{ans}}$, with their group-relative advantages $A_j^{\mathrm{ans}}$ normalized independently from the skill-update group. Both trajectory types are used to update the same policy. For trajectory type $b\in\{\mathrm{skill},\mathrm{ans}\}$, the token-level advantage is
\begin{equation}
A_{j,i}^{b}
=
\begin{cases}
\widetilde{A}_{j,i}^{\mathrm{skill}},
& b=\mathrm{skill},\\
A_j^{\mathrm{ans}},
& b=\mathrm{ans},
\end{cases}
\label{eq:training_advantage}
\end{equation}
where the response advantage $A_j^{\mathrm{ans}}$ is computed by independently normalizing $R_{\mathrm{ans}}$ in the same manner as Eq.~\ref{eq:skill_advantage}. For token $i$ in trajectory $j$, the policy ratio is defined as
\begin{equation}
r_{j,i}^{b}(\theta)
=
\frac{
\pi_{\theta}(o_{j,i}^{b}\mid o_{j,<i}^{b})
}{
\pi_{\theta_{\mathrm{old}}}
(o_{j,i}^{b}\mid o_{j,<i}^{b})
}.
\label{eq:policy_ratio}
\end{equation}
The conditioning context is omitted for clarity. The GRPO loss for each trajectory type is
\begin{equation}
\begin{aligned}
\mathcal{L}_{b}
=
-\mathbb{E}_{j,i}\Big[
\min\Big(
&r_{j,i}^{b}A_{j,i}^{b},
\operatorname{clip}
\bigl(r_{j,i}^{b},1-\epsilon_c,\\[-2pt]
&1+\epsilon_c\bigr)A_{j,i}^{b}
\Big)
\Big]
+\beta\mathcal{L}_{\mathrm{KL}}^{b}.
\end{aligned}
\label{eq:grpo_loss}
\end{equation}
The final objective combines the two losses:
\begin{equation}
\mathcal{L}
=
\mathcal{L}_{\mathrm{skill}}
+
\mathcal{L}_{\mathrm{ans}}.
\label{eq:joint_grpo}
\end{equation}
Thus, skill-update and response trajectories use separate rewards and independently normalized advantages while jointly updating the same policy. At inference time, the trained policy updates personalized skills from user feedback and applies the retained rules to subsequent requests without reward computation or group sampling.

\section{Experiments}
\subsection{Experiment Setup}
\paragraph{Tasks and Datasets}
We evaluate our method across five writing tasks. Topic Writing, Yelp, and Email are used for training. For evaluation, we use disjoint held-out examples from these three tasks as the in-domain test set, and use Product Review and Abstract, which are not seen during training, as out-of-domain test tasks. Dataset statistics and train/test splits are summarized in Table~\ref{tab:benchmark_tasks}.

\paragraph{Baselines}
We compare \textsc{SkillGrain} with both inference-only and training-based baselines. Among inference-only methods, \textsc{No Skill} answers queries without any explicit skill, \textsc{General Skill} uses a fixed general-purpose skill, and \textsc{Adaptive Skill} updates the skill from user feedback at inference time. \textsc{No History} uses the fixed general-purpose skill without access to the user's interaction history. \textsc{User Profile} summarizes the user's interaction history into a profile for generation, \textsc{RAG-History} retrieves the most relevant historical interaction as additional context, and \textsc{Feedback} directly uses the observed user feedback to guide the subsequent response. \textsc{Atom-Prompt} applies the same atomic decomposition and transferability protocol as \textsc{SkillGrain} at inference time without parameter optimization. For training-based methods, \textsc{SFT} learns from teacher-generated skill-update and response trajectories, while \textsc{STaR} performs self-training on high-reward trajectories generated by the model itself. \textsc{Answer-Only RL} applies GRPO only to skill-conditioned response generation; \textsc{Outcome-RL} jointly optimizes skill updating and response generation, using downstream response quality to evaluate skill updates; and \textsc{Rule-RL} directly scores updated skills with the same rubric-based reward used by \textsc{SkillGrain}. More  details are provided in Appendix~\ref{appendix:train_details}.

\paragraph{Implementation}
We use models from the Qwen2.5~\cite{qwen2025qwen25technicalreport} and Qwen3~\cite{yang2025qwen3} families as our base models, including Qwen2.5-3B-Instruct, Qwen3-1.7B, and Qwen3-4B. For all Qwen3 models, we disable the thinking mode throughout training and evaluation. For RL-based methods, we use a learning rate of $1\times10^{-6}$, a batch size of 14, and a KL coefficient of $0.001$. During training, the rollout temperature is set to $1.3$, with maximum generation lengths of 512 tokens for skill updating and 1,024 tokens for response generation. Further implementation details are provided in Appendix~\ref{appendix:train_details}.

\paragraph{Evaluation}
For evaluation, we use a sampling temperature of $0.7$, top-$p$ of $0.8$, and a maximum response length of 1,024 tokens. We repeat each evaluation three times with different random seeds and report the average performance across runs. Dimension-wise and atomic transferability results are provided in Appendices~\ref{appendix:dimension_results} and~\ref{appendix:atomic_transferability}, respectively.

\begin{table*}[t]
\centering
\small
\setlength{\tabcolsep}{4.5pt}
\renewcommand{\arraystretch}{0.8}

\begin{tabular}{l|ccc|cc|c}
\toprule
\multirow{2}{*}{\textbf{Methods}}
& \multicolumn{3}{c|}{\textbf{In-Domain}}
& \multicolumn{2}{c|}{\textbf{Out-of-Domain}}
& \multirow{2}{*}{\textbf{Average}} \\
& Topic Writing & Yelp & Email & Product Review & Abstract & \\
\midrule

\multicolumn{7}{l}{\textbf{Qwen2.5-3B-Instruct based}} \\
\midrule
No Skill
& 73.61 & 53.23 & 68.71 & 75.88 & 80.47 & 70.38 \\
No History
& 71.76 & 72.96 & 81.57 & 71.06 & 75.91 & 74.65 \\
General Skill
& 71.09 & 63.52 & 74.36 & 78.11 & 81.99 & 73.82 \\
Adaptive Skill
& 71.97 & 52.59 & 70.10 & \textbf{79.15} & \underline{82.45} & 71.25 \\
User Profile
& 68.95 & 53.75 & 59.20 & 74.96 & 82.37 & 67.85 \\
RAG-History
& 71.21 & 59.85 & 68.22 & 75.26 & 80.95 & 71.09 \\
Feedback
& 71.63 & 56.87 & 64.89 & 75.94 & 82.33 & 70.33 \\
Atom-Prompt
& 71.20 & 55.10 & 65.64 & 74.97 & 80.09 & 69.40 \\
SFT
& 72.21 & 79.37 & 80.20 & 76.31 & 81.69 & 77.96 \\
STaR
& 74.54 & 54.96 & 71.59 & \underline{78.45} & \textbf{82.82} & 72.47 \\
Answer-Only RL
& 73.76 & \underline{83.14} & 87.05 & 76.15 & 79.72 & 79.97 \\
Outcome-RL
& \underline{76.76} & 81.43 & 85.82 & 77.36 & 77.89 & 79.85 \\
Rule-RL
& \textbf{76.90} & 82.18 & \underline{87.71} & 76.56 & 78.39 & \underline{80.35} \\
\rowcolor{yellow!6}
SkillGrain (Ours)
& \textbf{76.90} & \textbf{85.11} & \textbf{87.77}
& 78.29 & 81.75 & \textbf{81.96} \\

\midrule
\multicolumn{7}{l}{\textbf{Qwen3-1.7B based}} \\
\midrule
No Skill
& 70.07 & 70.08 & 68.73 & 74.14 & 82.52 & 73.11 \\
No History
& 67.02 & 73.17 & 76.60 & 68.77 & 76.80 & 72.47 \\
General Skill
& 69.05 & 72.38 & 73.77 & 74.79 & 82.58 & 74.51 \\
Adaptive Skill
& 69.02 & 66.42 & 67.98 & 77.01 & 82.03 & 72.49 \\
User Profile
& 68.96 & 64.79 & 64.34 & 74.54 & 80.84 & 70.69 \\
RAG-History
& 66.68 & 64.40 & 67.99 & 73.93 & 77.79 & 70.15 \\
Feedback
& 68.42 & 64.92 & 66.58 & 74.16 & 81.92 & 71.20 \\
Atom-Prompt
& 68.29 & 62.38 & 64.67 & 76.19 & 83.01 & 70.90 \\
SFT
& 73.33 & 81.70 & 77.77 & 76.26 & 84.69 & 78.75 \\
STaR
& 70.06 & 68.59 & 69.17 & 76.58 & 83.04 & 73.48 \\
Answer-Only RL
& 77.65 & 85.25 & \underline{82.93} & 80.49 & 84.68 & 82.20 \\
Outcome-RL
& \underline{77.79} & 85.37 & 82.33 & \textbf{83.00} & 85.97 & \underline{82.89} \\
Rule-RL
& 75.95 & \underline{86.63} & 81.27 & 80.48 & \textbf{86.55} & 82.17 \\
\rowcolor{yellow!6}
SkillGrain (Ours)
& \textbf{79.12} & \textbf{89.00} & \textbf{84.25}
& \underline{81.65} & \underline{86.10} & \textbf{84.02} \\

\midrule
\multicolumn{7}{l}{\textbf{Qwen3-4B based}} \\
\midrule
No Skill
& 71.13 & 73.09 & 77.52 & 78.75 & 84.61 & 77.02 \\
No History
& 68.02 & 59.51 & 76.84 & 71.01 & 74.99 & 70.07 \\
General Skill
& 70.91 & 76.87 & 77.82 & 78.37 & 81.90 & 77.17 \\
Adaptive Skill
& 71.31 & 74.03 & 75.35 & 79.43 & 83.35 & 76.69 \\
User Profile
& 67.20 & 63.47 & 72.79 & 79.87 & 84.61 & 73.58 \\
RAG-History
& 68.39 & 63.45 & 69.34 & 76.86 & 77.86 & 71.18 \\
Feedback
& 67.38 & 66.42 & 66.10 & 78.09 & 81.90 & 71.97 \\
Atom-Prompt
& 73.89 & 76.22 & 74.97 & 78.48 & 82.67 & 77.24 \\
SFT
& 75.74 & 82.35 & 79.79 & 80.49 & 84.53 & 80.58 \\
STaR
& 72.94 & 76.07 & 76.72 & 79.84 & 83.31 & 77.77 \\
Answer-Only RL
& \underline{83.17} & 86.04 & 88.14 & 84.10 & 85.90 & 85.47 \\
Outcome-RL
& 81.03 & 89.09 & \underline{88.18} & 82.36 & 83.69 & 84.87 \\
Rule-RL
& 81.53 & \underline{89.88} & 87.58
& \underline{84.75} & \underline{86.11} & \underline{85.97} \\
\rowcolor{yellow!6}
SkillGrain (Ours)
& \textbf{84.59} & \textbf{90.06} & \textbf{88.35}
& \textbf{85.12} & \textbf{88.36} & \textbf{87.29} \\

\bottomrule
\end{tabular}

\caption{
Performance across five personalization tasks using three backbone models. The best and second-best results within each model block are shown in
bold and underlined, respectively.
}
\label{tab:main-results}
\end{table*}

\subsection{Main Results}
We evaluate all methods across five personalization tasks, with results reported in Table~\ref{tab:main-results}. Among inference-only methods, \textsc{No History}, \textsc{General Skill}, and \textsc{Atom-Prompt} achieve the highest average scores on Qwen2.5-3B-Instruct, Qwen3-1.7B, and Qwen3-4B, respectively, suggesting that inference-time personalization depends on the backbone. Notably, \textsc{Atom-Prompt}, which applies the same atomic decomposition and transferability protocol as \textsc{SkillGrain} without training, does not consistently outperform \textsc{No Skill}, indicating that structured prompting alone cannot explain our gains. Among training-based methods, SFT consistently outperforms STaR, but both trail the RL-based approaches. Rule-RL further outperforms Outcome-RL on two of the three backbones, supporting the value of directly evaluating skill quality rather than relying solely on downstream response rewards. Overall, \textsc{SkillGrain} achieves the highest average performance across all three backbones, surpassing the strongest baseline by 1.61, 1.13, and 1.32 points on Qwen2.5-3B-Instruct, Qwen3-1.7B, and Qwen3-4B, respectively, demonstrating the effectiveness of our approach.

\subsection{Ablation Study}
To assess the contribution of each component in \textsc{SkillGrain}, we conduct a series of ablation studies. \textit{w/ outcome-based reward} replaces the direct rubric-based skill reward with the reward of the subsequent response generated using the candidate skill; \textit{w/o atomization} replaces atomic skill induction with holistic KEEP/UPDATE updates while retaining matched-span local credit; \textit{w/o advantage allocation} removes the rubric-guided local advantage assignment; and \textit{w/o skill-update RL} disables RL optimization of skill-update trajectories and trains only skill-conditioned response generation. As shown in Table~\ref{tab:ablation}, removing atomization or skill-update RL causes the largest performance drops, while other ablations also reduce performance, supporting the contribution of each component.

\begin{table*}[t]
\centering
\small
\setlength{\tabcolsep}{4.5pt}
\renewcommand{\arraystretch}{0.9}
\begin{tabular}{l|ccccc|c}
\toprule
\textbf{Backbone / Method}
& \textbf{Topic} & \textbf{Yelp} & \textbf{Email}
& \textbf{Product} & \textbf{Abstract}
& \textbf{Average} \\
\midrule
\multicolumn{7}{l}{\textbf{Qwen2.5-3B-Instruct based}} \\
\midrule
\rowcolor{yellow!6}
SkillGrain
& 76.90 & \textbf{85.11} & \underline{87.77}
& \textbf{78.29} & 81.75 & \textbf{81.96} \\
\quad w/ outcome-based reward
& 76.75 & 82.94 & 86.24
& 76.84 & \textbf{82.12} & 80.99 \\
\quad w/o atomization
& \underline{77.57} & 79.89 & 87.11
& 73.56 & 76.46 & 78.93 \\
\quad w/o advantage allocation
& 77.12 & \underline{84.90} & \textbf{88.25}
& \underline{77.76} & \underline{81.45} & \underline{81.90} \\
\quad w/o skill update
& \textbf{77.61} & 78.74 & 84.77
& 75.31 & 80.16 & 79.33 \\
\midrule
\multicolumn{7}{l}{\textbf{Qwen3-1.7B based}} \\
\midrule
\rowcolor{yellow!6}
SkillGrain
& \textbf{79.12} & \underline{89.00} & \textbf{84.25}
& \textbf{81.65} & \textbf{86.10} & \textbf{84.02} \\
\quad w/ outcome-based reward
& \underline{77.91} & \textbf{89.90} & 84.02
& \underline{81.20} & 85.69 & \underline{83.75} \\
\quad w/o atomization
& 75.53 & 88.20 & 83.23
& 78.72 & 85.40 & 82.22 \\
\quad w/o advantage allocation
& 77.36 & 88.09 & \underline{84.13}
& 80.95 & \underline{85.90} & 83.29 \\
\quad w/o skill update
& 76.88 & 86.02 & 83.57
& 76.87 & 84.42 & 81.56 \\
\midrule
\multicolumn{7}{l}{\textbf{Qwen3-4B based}} \\
\midrule
\rowcolor{yellow!6}
SkillGrain
& \textbf{84.59} & 90.06 & 88.35
& \textbf{85.12} & \textbf{88.36} & \textbf{87.29} \\
\quad w/ outcome-based reward
& 83.01 & \textbf{91.57} & \textbf{88.95}
& 83.41 & 87.01 & \underline{86.80} \\
\quad w/o atomization
& 80.69 & \underline{90.84} & \underline{88.78}
& 80.45 & 85.35 & 85.23 \\
\quad w/o advantage allocation
& \underline{83.68} & 90.54 & 88.03
& \underline{84.14} & \underline{87.32} & 86.75 \\
\quad w/o skill update
& 81.67 & 86.71 & 85.54
& 83.31 & 83.50 & 84.15 \\
\bottomrule
\end{tabular}
\caption{Ablation results of SkillGrain. Best and second-best results under each backbone are shown in \textbf{bold} and \underline{underline}, respectively.}
\label{tab:ablation}
\end{table*}

\subsection{Effect of Adaptation Rounds}
\begin{figure*}[t]
    \centering

    \begin{minipage}[t]{0.24\textwidth}
        \centering
        \includegraphics[width=\linewidth]{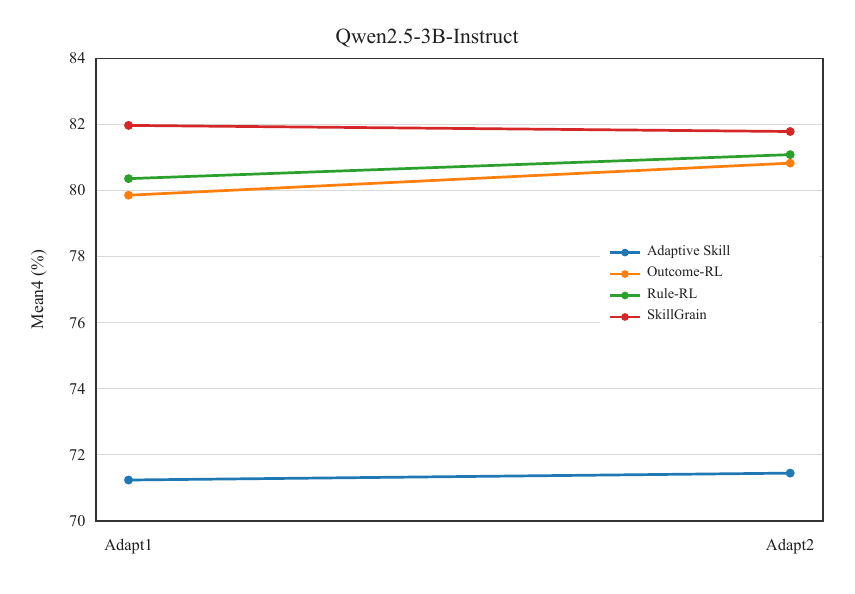}
    \end{minipage}
    \hfill
    \begin{minipage}[t]{0.24\textwidth}
        \centering
        \includegraphics[width=\linewidth]{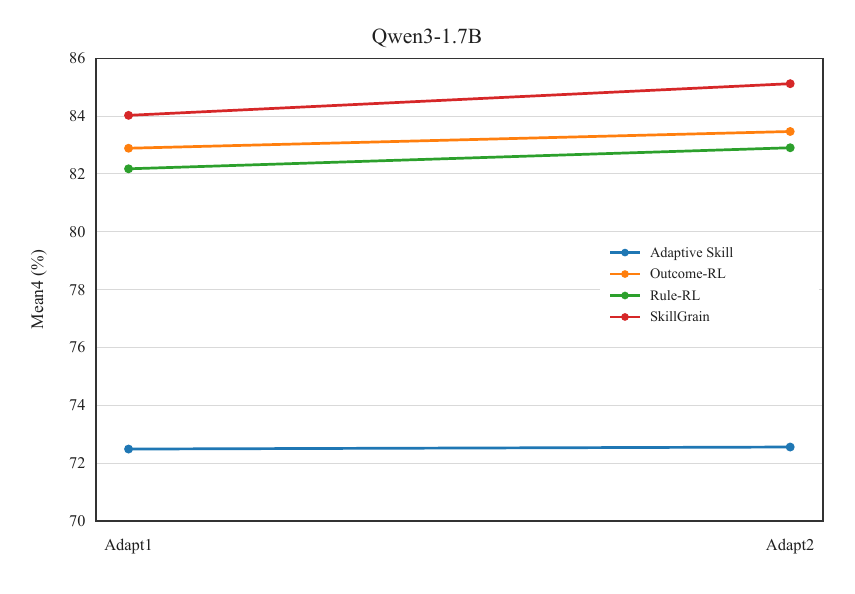}
    \end{minipage}
    \hfill
    \begin{minipage}[t]{0.24\textwidth}
        \centering
        \includegraphics[width=\linewidth]{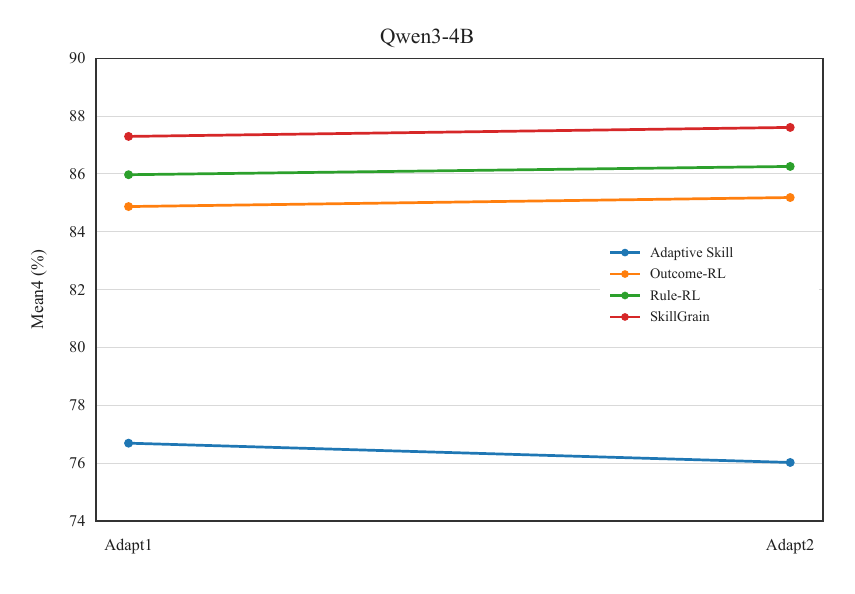}
    \end{minipage}
    \hfill
    \begin{minipage}[t]{0.24\textwidth}
        \centering
        \includegraphics[width=\linewidth]{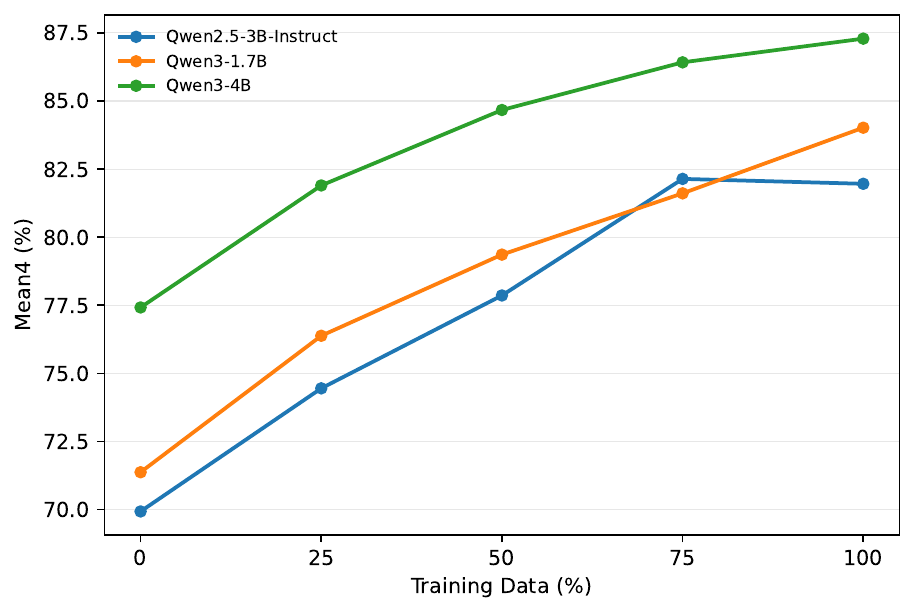}
    \end{minipage}

    \caption{
    Further analysis of SkillGrain.
    \textbf{(Left three):} Performance of trained methods across two rounds of feedback adaptation on Qwen2.5-3B-Instruct, Qwen3-1.7B, and Qwen3-4B, respectively.
    \textbf{(Right):} Performance of SkillGrain with different proportions of training data.
    Scores are Mean4 (\%) averaged over three evaluation seeds.
    }
    \label{fig:further_analysis}
\end{figure*}

To examine the effect of repeated user interactions, we compare \textsc{SkillGrain} against several competitive baselines after the first and second adaptation rounds. As shown in Figure~\ref{fig:further_analysis} (Left three), most methods further improve as additional feedback becomes available, indicating that repeated interaction provides useful signals for personalized adaptation. Importantly, \textsc{SkillGrain} consistently maintains the best performance across all three backbones after the second adaptation round and, in most cases, further enlarges its advantage over competing methods. These results demonstrate that \textsc{SkillGrain} effectively refines personalized skills through iterative interactions.

\subsection{Effect of Training Data Scale}
We train \textsc{SkillGrain} with different proportions of the available training data. As shown in Figure~\ref{fig:further_analysis} (Right), performance generally improves with increasing data scale, while the gains gradually saturate as the full training set is approached. This demonstrates that \textsc{SkillGrain} effectively benefits from additional training data.

\begin{figure}[t]
    \centering
    \includegraphics[width=0.8\linewidth]{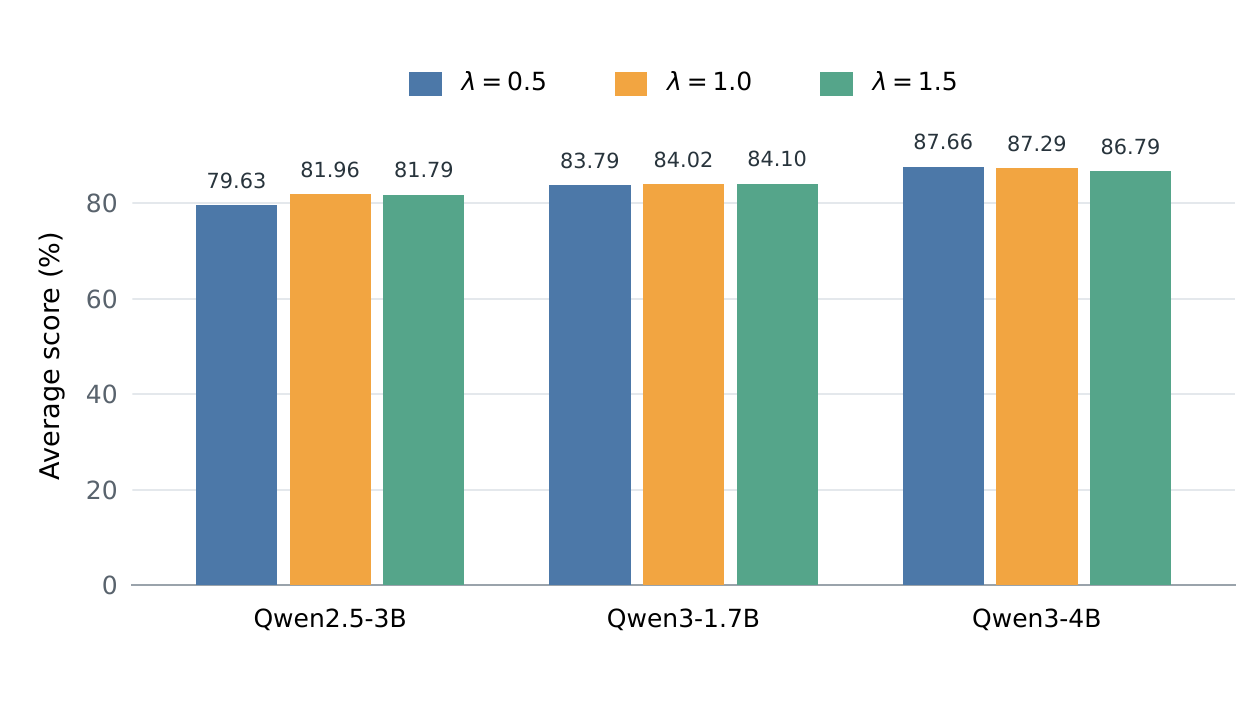}
    \caption{Sensitivity analysis of $\lambda$.}
    \label{fig:lambda_sensitivity}
\end{figure}

\subsection{Transfer to Alternative Rubrics}

To assess generalization beyond the training rubrics, we evaluate four unseen writing preferences: concise opening sentences, explicit causal connectives, continuous prose without headings or lists, and avoidance of hedging expressions. We construct 40 held-out request pairs and evaluate rule adherence using deterministic checks without further training. As shown in Table~\ref{tab:alternative_rubrics}, \textsc{SkillGrain} achieves the best or tied-best average performance across all three backbones, providing preliminary evidence of its ability to learn and reuse transferable rules beyond the training rubrics.

\begin{table}[t]
\centering
\small
\setlength{\tabcolsep}{5pt}
\renewcommand{\arraystretch}{1.05}
\begin{tabular}{lccc}
\toprule
\textbf{Method}
& \shortstack{\textbf{Qwen2.5}\\\textbf{3B-Instruct}}
& \shortstack{\textbf{Qwen3}\\\textbf{1.7B}}
& \shortstack{\textbf{Qwen3}\\\textbf{4B}} \\
\midrule
\textsc{Outcome-RL} & 72.5 & 65.0 & 77.5 \\
\textsc{Rule-RL} & 80.0 & \textbf{72.5} & 77.5 \\
\textsc{SkillGrain} & \textbf{82.5} & \textbf{72.5} & \textbf{82.5} \\
\bottomrule
\end{tabular}
\caption{Average rule adherence (\%) across four alternative writing preferences.}
\label{tab:alternative_rubrics}
\end{table}

\subsection{Sensitivity to Local-Credit Weight}

We examine SkillGrain's sensitivity to the local-credit weight $\lambda$ in Eq.~\ref{eq:local_advantage}, which scales credit for locally verified rule content and transferability decisions. We test $\lambda \in \{0.5, 1.0, 1.5\}$ under the same protocol. Figure~\ref{fig:lambda_sensitivity} shows modest performance changes across this range, with no setting consistently best across backbones. The default $\lambda=1.0$ remains competitive, indicating robustness to moderate variations in this weight.

\section{Conclusion}
We study feedback-driven personalized skill learning for writing tasks and propose \textsc{SkillGrain} to handle feedback containing both transferable and task-specific information. \textsc{SkillGrain} decomposes feedback into atomic rules, learns their transferability, and uses rubric-based rewards with fine-grained advantage assignment to jointly optimize skill updating and utilization. We further construct a benchmark spanning five writing tasks, and experiments across multiple backbones demonstrate the effectiveness of our approach.

\section*{Limitations}
Our study has several limitations. First, some user profiles in our benchmark are synthesized by LLMs, which may not fully capture the diversity and complexity of real user preferences. Besides, due to computational constraints, our experiments mainly focus on models in the 1.7B--4B range and writing-oriented tasks. Evaluating \textsc{SkillGrain} on larger models, real-world users, and more complex task domains remains an important direction for future work.

\section*{Ethical Considerations}
Personalized skills may retain or reproduce user-specific details from
interaction histories, and source corpora may contain identifiable or
objectionable text. Incorrectly transferring a preference or a factual
detail to a new request may also produce misleading personalized
responses. Our non-leakage rubric measures selected forms of factual
carryover; it does not establish privacy protection or content safety.
These risks require separate review before releasing derived data or
deploying the method with real users.

\section*{Acknowledgments}
This work was supported by the National Key Research and Development Program of China (Grant No. 2025YFF1505704), the National Science and Technology Major Project (Grant No. 2022ZD0117402), the National Natural Science Foundation of China (Grant No. 62441617), and the Beijing Advanced Innovation Center for Future Blockchain and Privacy Computing.


\bibliography{custom}

\appendix


\section{Benchmark Construction and Evaluation Details}
\label{appendix:benchmark}

\subsection{Benchmark Construction}
\label{appendix:benchmark_details}

\begin{table}[t]
\centering
\small
\setlength{\tabcolsep}{3pt}
\renewcommand{\arraystretch}{1.08}
\begin{tabular}{@{}p{0.22\linewidth}p{0.46\linewidth}rr@{}}
\toprule
\textbf{Task} & \textbf{Writing Scenario} & \textbf{Train} & \textbf{Test} \\
\midrule
Topic Writing
& Generate a personalized post from a topic summary.
& 498 & 200 \\

Email
& Continue an email from a given prefix in the user's style.
& 500 & 200 \\

Yelp
& Generate a personalized business review from metadata and a rating.
& 500 & 200 \\

Product Review
& Generate a personalized review from product information and a rating.
& -- & 198 \\

Abstract
& Generate an author-style abstract from a title and keywords.
& -- & 200 \\
\midrule
\textbf{Total} & & \textbf{1,498} & \textbf{998} \\
\bottomrule
\end{tabular}
\caption{Personalized writing tasks and curated split sizes.}
\label{tab:benchmark_tasks}
\end{table}

We construct five personalized writing tasks from publicly available datasets.
\textsc{Topic Writing}, \textsc{Product Review}, and \textsc{Abstract} are
adapted from LongLaMP~\cite{kumar2024longlamp}; \textsc{Email} is constructed
from the Enron Personalization Validation Set~\cite{fowler2015effects}; and
\textsc{Yelp} is derived from the Yelp Open Dataset~\cite{yelp_open_dataset}.
Table~\ref{tab:benchmark_tasks} summarizes the writing scenarios and curated
train and test split sizes.

Examples are grouped by user or author identity. For each benchmark instance,
we select five historical writings from the same user and construct additional
interaction queries with distinct task content. GPT-5.6 Terra is used to
synthesize interaction queries based on the user history and task
specification. We remove duplicate and near-duplicate examples and further
apply automatic validation and LLM-based consistency checks to filter
malformed or inconsistent instances.

The joint training pool contains 1,498 instances from \textsc{Email},
\textsc{Topic Writing}, and \textsc{Yelp}. The test split contains 998
instances across all five tasks. Consequently, \textsc{Abstract} and
\textsc{Product Review} provide evaluation on task types not observed in the
joint training pool. Two topic-writing training records and two
product-review test records are excluded because fewer than five distinct
historical writings remain after deduplication.

To support rubric-based evaluation and feedback construction, we derive
user-level writing statistics and grounded task-relevant entities or keywords
from the source examples. These annotations are used only for benchmark
construction, reward computation, and feedback generation, and are never
directly exposed to the evaluated policy.

\paragraph{Licenses and artifact release.}
The LongLaMP dataset card does not specify a license for redistributing the dataset. The Enron Personalization Validation Set is released for benchmarking purposes and does not state an explicit permission to redistribute its email text. The Yelp Open Dataset is governed by the Yelp Dataset Terms of Use, which permit academic use but restrict sharing the data with third parties. Our public artifact contains implementation code and configurations. It does not contain source or derived benchmark texts, user-level records, entity annotations, generated outputs, or model checkpoints.

\subsection{Rubric Scoring}
\label{appendix:rubric_scoring}

We evaluate generated responses along four dimensions:
length preference (\textsc{Len}),
first-person usage (\textsc{FP}),
expression preference (\textsc{Expr}),
and non-leakage (\textsc{NL}).
All dimensions are normalized to $[0,1]$.

\paragraph{Interval-Based Scoring.}
Both \textsc{Len} and \textsc{FP} use a piecewise-linear interval function.
Given a preferred interval $[l,h]$ and outer bounds $[o_l,o_h]$, we define
\begin{equation}
\phi(v)=
\begin{cases}
1, & l\leq v\leq h,\\
0, & v\leq o_l \text{ or } v\geq o_h,\\
\dfrac{v-o_l}{l-o_l}, & o_l<v<l,\\
\dfrac{o_h-v}{o_h-h}, & h<v<o_h.
\end{cases}
\label{eq:interval_score}
\end{equation}
Values within the preferred interval receive full credit, while deviations are penalized linearly.

\paragraph{Length Preference (\textsc{Len}).}
Let $m_u$ denote the median word count of the five historical writings.
We define the preferred lower bound as
\begin{equation}
l_{\mathrm{len}}=0.7m_u,
\end{equation}
and the preferred upper bound as
\begin{equation}
h_{\mathrm{len}}=1.3m_u.
\end{equation}
The corresponding outer bounds are
\begin{equation}
o_{\mathrm{len}}^{l}=0.5l_{\mathrm{len}},
\end{equation}
and
\begin{equation}
o_{\mathrm{len}}^{h}=1.5h_{\mathrm{len}}.
\end{equation}
For a generated response $y$, the length score is
\begin{equation}
r_{\mathrm{len}}(y)
=
\phi
\left(
w(y);
l_{\mathrm{len}},
h_{\mathrm{len}},
o_{\mathrm{len}}^{l},
o_{\mathrm{len}}^{h}
\right),
\label{eq:length_score}
\end{equation}
where $w(y)$ denotes the response word count.

\paragraph{First-Person Usage (\textsc{FP}).}
We compute the number of first-person tokens per 100 words over the concatenated historical writings.
The first-person lexicon includes
\textit{I, me, my, mine, we, us, our, ours}
and their common contractions.
Let $q_u$ denote the historical first-person rate.
We first define
\begin{equation}
\tau_u=\max(2.0,0.6q_u).
\end{equation}
The preferred lower bound is
\begin{equation}
l_{\mathrm{fp}}
=
\max(0,q_u-\tau_u),
\end{equation}
and the preferred upper bound is
\begin{equation}
h_{\mathrm{fp}}
=
q_u+\tau_u.
\end{equation}
The outer bounds are
\begin{equation}
o_{\mathrm{fp}}^{l}
=
\max(0,l_{\mathrm{fp}}-\tau_u),
\end{equation}
and
\begin{equation}
o_{\mathrm{fp}}^{h}
=
h_{\mathrm{fp}}+\tau_u.
\end{equation}
The first-person score is then
\begin{equation}
r_{\mathrm{fp}}(y)
=
\phi
\left(
q(y);
l_{\mathrm{fp}},
h_{\mathrm{fp}},
o_{\mathrm{fp}}^{l},
o_{\mathrm{fp}}^{h}
\right),
\label{eq:first_person_score}
\end{equation}
where $q(y)$ denotes the first-person rate of response $y$.

\paragraph{Expression Preference (\textsc{Expr}).}
We consider six automatically measurable expression features:
contractions, exclamation marks, question marks, parenthetical expressions, all-caps emphasis, and average sentence length.
For each user, we select one feature that is stable across the historical writings and sufficiently distinctive from the population-level tendency.

Let $\mu_u$ denote the user-level median of the selected feature and $\mu_p$ its population median.
We define the feature scale as
\begin{equation}
s_u=
\max\left\{
|\mu_u-\mu_p|,
\operatorname{Range}_u,
0.25
\right\}.
\label{eq:expr_scale}
\end{equation}
The no-penalty tolerance is
\begin{equation}
\delta_u=
\max\left\{
\operatorname{MAD}_u,
0.05s_u
\right\}.
\label{eq:expr_tolerance}
\end{equation}
For a response $y$, we first compute its deviation from the user preference:
\begin{equation}
d_u(y)=
\max\left\{
0,
|e(y)-\mu_u|-\delta_u
\right\},
\label{eq:expr_deviation}
\end{equation}
where $e(y)$ denotes the selected feature value of $y$.
The expression score is then
\begin{equation}
r_{\mathrm{expr}}(y)
=
\max\left\{
0,
1-\frac{d_u(y)}{s_u}
\right\}.
\label{eq:expression_score}
\end{equation}

\paragraph{Non-Leakage (\textsc{NL}).}
To detect inappropriate information transfer across interactions, we construct a leakage candidate set from task-relevant entities and keywords appearing in historical or previous interactions.
Terms supported by the current request are excluded from this set.
Let $\mathcal{L}_{u,t}(y)$ denote the set of distinct leakage candidates detected in response $y$.
We define
\begin{equation}
r_{\mathrm{nl}}(y)
=
2^{-|\mathcal{L}_{u,t}(y)|}.
\label{eq:nonleak_score}
\end{equation}
Thus, zero, one, two, and three distinct leakage hits yield scores of
$1$, $0.5$, $0.25$, and $0.125$, respectively.

\paragraph{Overall Score.}
The final response reward is the equally weighted average of the four dimensions:
\begin{equation}
R_{\mathrm{ans}}(y)
=
\frac{1}{4}
\left(
r_{\mathrm{len}}
+r_{\mathrm{fp}}
+r_{\mathrm{expr}}
+r_{\mathrm{nl}}
\right).
\label{eq:mean4_appendix}
\end{equation}

\subsection{Rubric-Grounded Feedback Construction}
\label{appendix:feedback_details}
Directly prompting an LLM to generate user feedback can introduce substantial variability and make the supervision difficult to control. To obtain reliable yet realistic feedback, we first use deterministic rubric rules to identify revision targets from the model response, including reusable preference corrections such as length or stylistic adjustments, as well as interaction-specific requirements such as missing current-request information or irrelevant content carried over from previous interactions. This design allows the feedback to reflect the mixture of transferable preferences and local requirements that may naturally occur in user interactions. We then use a fixed Qwen3-4B model with thinking disabled and deterministic decoding to verbalize these structured revision targets into concise natural-language feedback. The verbalizer is instructed to preserve the original revision requirements without introducing additional preferences or facts.

\section{Experimental Details}
\label{appendix:exp_details}

\subsection{Fine-Grained Credit Assignment}
\label{appendix:credit_details}

For local credit assignment, we deterministically align each generated atomic rule with the structured revision targets retained during feedback construction. For preference-related rules, we map each target to a canonical direction (e.g., more/less concise or more/less first-person) and match the generated rule using predefined lexical and paraphrase patterns, while also checking for the opposite direction. For interaction-specific rules, we first match grounded entities or key phrases and then verify whether the associated directive expresses the expected action, such as including or removing the information. Ambiguous, conflicting, or unsupported matches are treated as abstentions and retain the original sequence-level advantage.

\paragraph{Omission penalty.}
Field-level credit only supervises atoms that are actually generated, so an entirely omitted transferable target would otherwise receive no local negative signal. Let $\mathcal{M}$ denote the set of confidently missing transferable targets and $\mathcal{T}^{+}$ the set of transferable targets in the feedback. We apply an omission penalty only when the unmatched transferable targets can be confidently identified:
\[
c_{\mathrm{omit}}
=
-\frac{|\mathcal{M}|}{|\mathcal{T}^{+}|}.
\]
The score is assigned to the sequence-closing token. When it is unclear whether an unmatched target has already been expressed by a generated atom, no omission penalty is applied.

\subsection{Training and Comparison Details}
\label{appendix:train_details}

Our framework is implemented on top of VeRL~\cite{sheng2024hybridflow}. All experiments were conducted on 4$\times$H100 GPUs with 80GB of memory, using Python 3.10 and PyTorch 2.6, with vLLM~\cite{kwon2023efficient} for rollout acceleration. Except for \textsc{No History}, all methods use five historical interactions from the same user as personalization context. \textsc{No History} uses the original base model and the same fixed general-purpose skill as \textsc{General Skill}, while omitting historical interactions from the response prompt. \textsc{Adaptive Skill}, \textsc{Outcome-RL}, and \textsc{Rule-RL} use the holistic keep-or-update skill protocol, whereas \textsc{SkillGrain} generates atomic rules with transferability predictions. \textsc{Atom-Prompt} uses the same atomic decomposition and transferability protocol as \textsc{SkillGrain} at inference time without parameter optimization. \textsc{Answer-Only RL} keeps the skill updater frozen and applies GRPO only to skill-conditioned response generation. \textsc{Outcome-RL} evaluates skill updates using downstream response rewards, while \textsc{Rule-RL} and \textsc{SkillGrain} use the same direct rubric-based skill reward. For \textsc{SFT}, Qwen3-8B generates teacher trajectories with deterministic decoding; \textsc{STaR} instead samples four skill-update and four response candidates from the target model and retains the highest-reward trajectories for supervised training.

All RL-based methods use GRPO with a global batch size of 14, a PPO mini-batch size of 14, and a micro-batch size of 1 per GPU. We train for one epoch with one PPO epoch per batch, a learning rate of $1\times10^{-6}$, weight decay of $0.01$, and a KL coefficient of $0.001$. Skill-update and response rollouts use $G=4$, temperature $1.3$, and top-$p=1.0$. The maximum prompt length is 16,384 tokens, while skill updates and responses are limited to 512 and 1,024 tokens, respectively. For \textsc{SkillGrain}, the local-credit weight is set to $\lambda=1.0$. SFT-based methods use AdamW with $(\beta_1,\beta_2)=(0.9,0.95)$, while RL-based methods follow the default VeRL AdamW setting $(0.9,0.999)$. For \textsc{SFT} and \textsc{STaR}, we use a batch size of 28, a micro-batch size of 1 per GPU, a maximum sequence length of 17,408 tokens, and the same learning rate of $1\times10^{-6}$ for one epoch.

\subsection{Dimension-wise Results}
\label{appendix:dimension_results}
Table~\ref{tab:dimension_results} reports the four rubric scores for the three RL methods. \textsc{SkillGrain} achieves the highest LEN score across all three backbones and improves NL over \textsc{Outcome-RL} throughout, reflecting better length alignment and reduced inappropriate factual carryover.

\begin{table}[t]
\centering
\small
\setlength{\tabcolsep}{4pt}
\renewcommand{\arraystretch}{1.05}
\begin{tabular}{lcccc}
\toprule
\textbf{Method} & \textbf{LEN} & \textbf{FP}
& \textbf{EXPR} & \textbf{NL} \\
\midrule
\multicolumn{5}{l}{\textit{Qwen2.5-3B-Instruct}} \\
\textsc{Outcome-RL} & 71.29 & 86.95 & 68.24 & 92.94 \\
\textsc{Rule-RL} & 70.83 & 87.39 & 67.98 & 95.23 \\
\textsc{SkillGrain}
& \textbf{74.62} & \textbf{88.05}
& \textbf{68.85} & \textbf{96.35} \\
\midrule
\multicolumn{5}{l}{\textit{Qwen3-1.7B}} \\
\textsc{Outcome-RL}
& 80.47 & \textbf{88.16} & 70.22 & 92.70 \\
\textsc{Rule-RL} & 79.25 & 85.76 & 69.55 & 94.15 \\
\textsc{SkillGrain}
& \textbf{82.65} & 87.58
& \textbf{70.67} & \textbf{95.22} \\
\midrule
\multicolumn{5}{l}{\textit{Qwen3-4B}} \\
\textsc{Outcome-RL} & 83.75 & 89.66 & 70.13 & 95.96 \\
\textsc{Rule-RL}
& 85.33 & 89.38 & \textbf{72.27} & \textbf{96.92} \\
\textsc{SkillGrain}
& \textbf{89.78} & \textbf{91.20} & 71.51 & 96.70 \\
\bottomrule
\end{tabular}
\caption{Dimension-wise rubric scores of the three RL
methods.}
\label{tab:dimension_results}
\end{table}

\subsection{Atomic Transferability Analysis}
\label{appendix:atomic_transferability}

To assess whether \textsc{SkillGrain} learns to distinguish transferable skill atoms from interaction-specific feedback, we examine the precision of its test-set atomic extraction and transferable-atom selection. As shown in Table~\ref{tab:atomic_transferability}, both remain high across backbones and evaluation domains, suggesting that the trained model can identify reusable preferences while limiting inappropriate transfer into persistent skills.

\begin{table}[t]
\centering
\footnotesize
\setlength{\tabcolsep}{3.5pt}
\renewcommand{\arraystretch}{1.08}
\begin{tabular}{@{}lcccc@{}}
\toprule
& \multicolumn{2}{c}{\textbf{Extraction}}
& \multicolumn{2}{c}{\textbf{Transferable}} \\
\cmidrule(lr){2-3}
\cmidrule(l){4-5}
\textbf{Backbone} & \textbf{ID} & \textbf{OOD}
& \textbf{ID} & \textbf{OOD} \\
\midrule
Qwen2.5-3B-Instruct & 89.31 & 87.04 & 84.27 & 79.60 \\
Qwen3-1.7B          & 82.63 & 77.71 & 86.24 & 81.41 \\
Qwen3-4B            & 94.54 & 90.30 & 89.81 & 81.80 \\
\bottomrule
\end{tabular}
\caption{Atomic extraction and transferable-atom selection precision (\%). ID and OOD denote in-domain and out-of-domain test sets, respectively.}
\label{tab:atomic_transferability}
\end{table}

\subsection{Evaluation with an Independent LLM Judge}
To complement the rubric-based evaluation, we use GPT-6 Luna as an independent judge to assess the three methods with Qwen2.5-3B-Instruct on 200 randomly sampled test instances (40 per task). Given user histories, interaction context, and anonymized responses, the judge evaluates length suitability, first-person usage, expression style, and avoidance of inappropriate factual carryover. As shown in Table~\ref{tab:llm_judge}, \textsc{SkillGrain} achieves the highest mean judge score among the three methods.

\begin{table}[t]
\centering
\small
\setlength{\tabcolsep}{6pt}
\renewcommand{\arraystretch}{1.05}
\begin{tabular}{lcc}
\toprule
\textbf{Method} & \textbf{Judge Score} \\
\midrule
\textsc{Outcome-RL} & 73.44 \\
\textsc{Rule-RL}    & 75.16 \\
\textsc{SkillGrain} & \textbf{76.44} \\
\bottomrule
\end{tabular}
\caption{Independent GPT-6 Luna evaluation with
Qwen2.5-3B-Instruct on the same 200 test instances.}
\label{tab:llm_judge}
\end{table}

\subsection{Comparison of Computational Cost}
To compare the training overhead of different methods, we report their training times using Qwen3-4B as the backbone model on four H100 GPUs, as summarized in Table~\ref{tab:time_cost}.

\begin{table}[t]
\centering
\small
\setlength{\tabcolsep}{8pt}
\renewcommand{\arraystretch}{1.1}
\begin{tabular}{lc}
\toprule
\textbf{Method} & \textbf{Training Time (h)} \\
\midrule
SFT & 2.0 \\
STaR & 2.0 \\
Answer-Only RL & 0.9 \\
Outcome-RL & 2.1 \\
Rule-RL & 1.7 \\
\rowcolor{yellow!6}
SkillGrain (Ours) & 1.8 \\
\bottomrule
\end{tabular}
\caption{Training time (hours) of methods requiring optimization with Qwen3-4B.}
\label{tab:time_cost}
\end{table}

\section{Training Algorithm}
\label{appendix:training_algorithm}

Algorithm~\ref{alg:skillgrain} summarizes the
training procedure of \textsc{SkillGrain}.

\begin{algorithm}[t]
\caption{Training Procedure of SkillGrain}
\label{alg:skillgrain}
\small
\begin{algorithmic}[1]
\Require Training data $\mathcal{D}_{\mathrm{train}}$,
policy $\pi_\theta$, group size $G$,
local-credit weight $\lambda$
\Ensure Trained policy $\pi_\theta$

\For{each training batch $\mathcal{B}$}
    \State $\theta_{\mathrm{old}} \gets \theta$
    \State $\mathcal{L}_{\mathcal{B}} \gets 0$

    \For{each $(c_t,\mathcal{V}_t,x_{t+1})\in\mathcal{B}$}
        \For{$j=1,\ldots,G$}
            \State Sample
            $\mathcal{A}^{(j)}
            =[(a_{j,k},z_{j,k})]_{k=1}^{K_j}
            \sim\pi_{\theta_{\mathrm{old}}}(\cdot\mid c_t)$
            \State $s^{(j)}
            \gets \operatorname{Compose}
            (s_t,\mathcal{A}^{(j)})$
            \State $R_j^{\mathrm{skill}}
            \gets R_{\mathrm{skill}}(s^{(j)})$
        \EndFor

        \State Compute $\mu_{\mathrm{skill}}$
        and $\sigma_{\mathrm{skill}}$ from
        $\{R_j^{\mathrm{skill}}\}_{j=1}^{G}$

        \For{$j=1,\ldots,G$}
            \State $A_j^{\mathrm{skill}}
            \gets
            \dfrac{R_j^{\mathrm{skill}}-\mu_{\mathrm{skill}}}
            {\sigma_{\mathrm{skill}}+\epsilon}$
            \State $\widetilde{A}_{j,i}^{\mathrm{skill}}
            \gets A_j^{\mathrm{skill}}$
            for all generated tokens $i$

            \For{each $(a_{j,k},z_{j,k})
            \in\mathcal{A}^{(j)}$}
                \State Align with targets in $\mathcal{V}_t$
                \State Obtain
                $c_{j,k}^{a},c_{j,k}^{z}$
                and token spans $T_{j,k}^{a},T_{j,k}^{z}$

                \If{$c_{j,k}^{a}\neq 0$}
                    \State $\widetilde{A}_{j,i}^{\mathrm{skill}}
                    \gets
                    \lambda c_{j,k}^{a}/|T_{j,k}^{a}|$,
                    $\forall i\in T_{j,k}^{a}$
                \EndIf

                \If{$c_{j,k}^{z}\neq 0$}
                    \State $\widetilde{A}_{j,i}^{\mathrm{skill}}
                    \gets
                    \lambda c_{j,k}^{z}/|T_{j,k}^{z}|$,
                    $\forall i\in T_{j,k}^{z}$
                \EndIf
            \EndFor

            \State Apply omission penalty to the
            closing token if required
        \EndFor

        \State $j^*\gets
        \arg\max_j R_j^{\mathrm{skill}}$
        \State $s^*\gets s^{(j^*)}$

        \For{$j=1,\ldots,G$}
            \State Sample
            $y_j\sim\pi_{\theta_{\mathrm{old}}}
            (\cdot\mid H_u,s^*,x_{t+1})$
            \State $R_j^{\mathrm{ans}}
            \gets R_{\mathrm{ans}}(y_j)$
        \EndFor

        \State Compute independently normalized
        response advantages $\{A_j^{\mathrm{ans}}\}_{j=1}^{G}$

        \State Compute $\mathcal{L}_{\mathrm{skill}}$
        using $\widetilde{A}_{j,i}^{\mathrm{skill}}$
        \State Compute $\mathcal{L}_{\mathrm{ans}}$
        using $A_j^{\mathrm{ans}}$
        \State $\mathcal{L}_{\mathcal{B}}\gets
        \mathcal{L}_{\mathcal{B}}
        +\mathcal{L}_{\mathrm{skill}}
        +\mathcal{L}_{\mathrm{ans}}$
    \EndFor

    \State $\theta\gets\theta-\eta\nabla_\theta
    \left(\mathcal{L}_{\mathcal{B}}/|\mathcal{B}|\right)$
\EndFor

\State \Return $\pi_\theta$
\end{algorithmic}
\end{algorithm}

\section{Use of Large Language Models}
Some parts of the text were refined with the assistance of large language models (LLMs). All content and responsibility for the work remain with the authors.

\section{Prompts}
This section lists the prompts used by the personalized-writing pipeline.
Except for the No History baseline, all methods use the same five historical examples, current request, actor model, and answering prompt template. The \textsc{No Skill} condition leaves the rule block empty; the \textsc{General Skill} and adaptation conditions differ only in which rules are supplied in that block. The judge is fixed and is used only to rewrite deterministic
editing notes into natural-language feedback; it does not compute rewards.

\noindent\textbf{Notation.}
\texttt{\{HISTORY\}} denotes five examples from the same user, and
\texttt{\{REQUEST\}}, \texttt{\{DRAFT\}}, and \texttt{\{FEEDBACK\}} denote
the current request, draft, and natural-language feedback, respectively.

\begin{figure}[!htbp]
\centering
\begin{minipage}{0.96\linewidth}
\begin{promptcolorbox}{Common Answer Prompt}{caseblue}
System:
 /no_think
You are a writing assistant. Return only the requested writing, with no analysis, labels, or markdown.

User:
Below are five examples from the same user. Complete the current request in that user's style. Use only facts supported by the current request; do not copy names or details from unrelated examples.

{HISTORY 1}
...
{HISTORY 5}

Writing skill rules:
r1: {RULE CONTENT}

Current request:
{REQUEST}
\end{promptcolorbox}
\end{minipage}
\caption{Common answering prompt shared by all methods.}
\label{fig:app-common-answer-prompt}
\end{figure}

\begin{figure}[!htbp]
\centering
\begin{minipage}{0.96\linewidth}
\begin{promptcolorbox}{Task-Specific General Skills}{casegray}
Abstract:
Write a concise, self-contained scientific abstract. State the problem, the proposed approach, the main result or contribution, and only claims supported by the current request. Use formal technical prose and do not invent experimental findings.

Email completion:
Continue naturally from the supplied prefix. Match the user's greeting, sentence length, punctuation, directness, formality, and sign-off habits. Address only the current email and do not repeat its prefix.

Product review:
Keep the review anchored to the current product and rating. Explain concrete strengths and limitations supported by the prompt, preserve the supplied sentiment, and write only the review in a natural consumer voice without invented specifications or experiences.

Topic writing:
Write a natural Reddit-style post that directly addresses the current topic, uses a coherent personal voice, and follows the user's usual level of detail and informality. Do not invent unsupported personal history.

Yelp:
Write only the Yelp review. Let the rating guide the stance, discuss only the current business and supplied experience, and match the user's usual brevity, punctuation, detail level, and tone without inventing visits or menu items.
\end{promptcolorbox}
\end{minipage}
\caption{Static task-specific skills used by the general-skill condition.}
\label{fig:app-general-skills}
\end{figure}

\begin{figure}[!htbp]
\centering
\begin{minipage}{0.96\linewidth}
\begin{promptcolorbox}{Rule Update Prompt}{casegold}
System:
 /no_think
You maintain a concise writing skill for one user.
Read the interaction and the user's natural feedback. Decide whether the feedback supports one durable, reusable writing rule.
A reusable rule describes writing style or process that is useful on a different topic. Do not store names, products, places, dates, topic-specific facts, or instructions that only repair the current answer.
If there is no supported reusable rule, choose KEEP. Otherwise add exactly one short rule.
Return JSON only, exactly one of:
{"decision":"KEEP","rule_updates":[],"new_rules":[]}
{"decision":"UPDATE","rule_updates":[],"new_rules":[{"content":"one reusable content-neutral rule"}]}

User:
User history:
{HISTORY}

Current request:
{REQUEST}

Assistant draft:
{DRAFT}

User feedback:
{FEEDBACK}

Return the KEEP or UPDATE JSON.
\end{promptcolorbox}
\end{minipage}
\caption{Rule-update prompt used by the outcome-reward and direct-rule
baselines.}
\label{fig:app-rule-update-prompt}
\end{figure}

\begin{figure}[!htbp]
\centering
\begin{minipage}{0.96\linewidth}
\begin{promptcolorbox}{Atom Decomposition Prompt (Ours)}{casegreen}
System:
 /no_think
Split the user's feedback into independently actionable requests.
For each request output only id (f1, f2, ...), content (a short concrete restatement), and transferable (true only when it applies to a different topic).
General writing preferences are transferable. Requests to insert or remove specific names, products, places, facts, numbers, examples, or technical terms are not transferable. Do not turn them into general instructions to incorporate terms, facts, examples, or references. Preserve each request's direction and qualifications. Include all feedback requests, including non-transferable ones.
The program will join the content of all transferable=true atoms with "; " to form the skill. Do not produce a separate summary, decision, or new_rule.
Example: Make this shorter. Include Paris and the Louvre naturally.
{"atoms":[{"id":"f1","content":"Make the writing shorter.","transferable":true},{"id":"f2","content":"Include Paris and the Louvre naturally.","transferable":false}]}
Return JSON only with the single top-level key atoms.

User:
User history:
{HISTORY}

Current request:
{REQUEST}

Assistant draft:
{DRAFT}

User feedback:
{FEEDBACK}
\end{promptcolorbox}
\end{minipage}
\caption{Atom-decomposition prompt used by our method. The program concatenates
only atoms whose \texttt{transferable} field is true.}
\label{fig:app-atom-prompt}
\end{figure}

\begin{figure}[!htbp]
\centering
\begin{minipage}{0.96\linewidth}
\begin{promptcolorbox}{Judge Feedback-Rewriting Prompt}{casegray}
System:
Rewrite the supplied editing notes as one natural paragraph of feedback addressed to the writer.
Your output must describe what should change; do not perform the rewrite and do not reproduce the draft.
Keep every requested change and its meaning, combine repetitive wording, and use no more than 80 words.
Do not add any preference, name, fact, example, explanation, or criticism absent from the notes.
Do not mention categories, scores, rules, or whether anything applies to future writing. Do not use a heading or bullet list.
Return only the feedback paragraph.

User:
Editing notes (the only source of feedback):
- {EDITING NOTE 1}
- {EDITING NOTE 2}
- ...
\end{promptcolorbox}
\end{minipage}
\caption{Prompt for the fixed judge model. Its input notes are constructed
deterministically from the four reward dimensions and entity diagnostics.}
\label{fig:app-judge-prompt}
\end{figure}

\FloatBarrier

\raggedbottom
\section{Case Study}
\label{app:case_study}

\begin{table}[t]
\centering
\small
\setlength{\tabcolsep}{4pt}
\renewcommand{\arraystretch}{1.08}
\begin{tabular}{@{}lccc@{}}
\toprule
\textbf{Task} & \textbf{Round 1} & \textbf{Round 2} & \textbf{Test} \\
\midrule
Abstract & 0.905 & 1.000 & 1.000 \\
Topic Writing & 0.750 & 1.000 & 0.903 \\
Email Completion & 0.528 & 0.992 & 0.889 \\
\bottomrule
\end{tabular}
\caption{Mean four-dimensional scores of the saved case-study trajectories.}
\label{tab:case-study-overview}
\end{table}

We qualitatively examine saved two-round trajectories produced by our
atom-based method. Table~\ref{tab:case-study-overview} reports the four-dimensional mean score before adaptation, after the first skill update, and on
the held-out test request. These examples are selected to illustrate the
interaction mechanism rather than to provide an aggregate comparison.

We detail the abstract-writing trajectory because it exposes both possible
skill-update outcomes: the first round adds a transferable preference, whereas
the second round correctly leaves the skill unchanged when the feedback is
entirely request-specific. The actor model is Qwen3-4B. Long history examples,
requests, and responses are shortened only for presentation.

\begin{casebox}{Context and User History}{casegray}
\small
The five abstracts exhibit formal technical prose, relatively long sentences,
and detailed descriptions. Representative passages discuss semantic
technologies, human input for data interlinking, gamification in citizen
science, human--machine mobility workflows, and Wikidata editor engagement.
\end{casebox}

\begin{casebox}{Round 1: Draft and Feedback}{caseblue}
\small
The request asks for an abstract about SeaFish, a game for collaborative image
annotation and interlinking. The generated draft describes human computation,
DBpedia and Flickr images, and Linked Data, but is shorter than the user's
historical preference.

\medskip
\textbf{Feedback.}
``The text is somewhat brief and could benefit from additional useful details.
Incorporate the terms `semantic content creation' and `data interlinking' more
naturally into the draft to enhance clarity and depth.''

\begin{center}
\scriptsize
\setlength{\tabcolsep}{2.5pt}
\begin{tabular}{@{}lccccc@{}}
\toprule
\textbf{Stage} & \textbf{LEN} & \textbf{FP} & \textbf{EXPR} & \textbf{NL} & \textbf{Mean} \\
\midrule
Round 1 & 0.620 & 1.000 & 1.000 & 1.000 & 0.905 \\
Round 2 & 1.000 & 1.000 & 1.000 & 1.000 & 1.000 \\
Test & 1.000 & 1.000 & 1.000 & 1.000 & 1.000 \\
\bottomrule
\end{tabular}
\end{center}
\end{casebox}

\begin{casebox}{Transferability Filtering and Skill Update}{casegreen}
\small
\begin{center}
\scriptsize
\setlength{\tabcolsep}{3pt}
\begin{tabular}{@{}p{0.68\linewidth}c@{}}
\toprule
\textbf{Feedback atom} & \textbf{Transferable} \\
\midrule
Make the text longer with additional useful details. & Yes \\
Incorporate semantic content creation naturally. & No \\
Incorporate data interlinking naturally. & No \\
\bottomrule
\end{tabular}
\end{center}

Only the first atom is retained, producing the reusable skill
\textit{Make the text longer with additional useful details.} The two named
terms describe content required by the current request and are therefore not
stored as user-level writing preferences.
\end{casebox}

\begin{casebox}{Round 2: Local Feedback Leads to KEEP}{casegreen}
\small
The next request concerns gamification and adaptive incentives for paid
microtask crowdsourcing. Conditioned on the retained skill, the response
expands the approach, experiments, and findings, obtaining full credit on all
four dimensions. The remaining feedback asks only for the term
``microtasks'' to be integrated more naturally. The method decomposes this
feedback into three non-transferable atoms and applies \textsc{Keep}; the
skill remains unchanged.
\end{casebox}

\begin{casebox}{Held-Out Test}{casegold}
\small
The test request asks for an abstract about the TheyBuyForYou platform and its
knowledge graph for public procurement. The response explains how open linked
data from multiple EU sources supports transparency and cross-border bidding.
It receives \textbf{1.000} for length, first-person usage, expression, and
non-leakage. All three target terms are covered, and no entity from the history
or adaptation requests is detected in the response.
\end{casebox}

\begin{casebox}{Interpretation}{casegray}
\small
The trajectory illustrates the intended transfer boundary. A general
preference for additional useful detail survives across unrelated abstract
topics, while request-specific terminology is used for local feedback but is
not written into persistent memory. The second-round \textsc{Keep} decision
also shows that an interaction need not modify the skill when its feedback
contains no transferable preference.
\end{casebox}

\FloatBarrier

\end{document}